\documentclass[journal]{IEEEtran}
\usepackage[pdftex]{graphicx}

\usepackage{enumitem}
\usepackage[utf8]{inputenc} 
\usepackage{balance}
\usepackage{cite}
\begin{document}
\title{Introduction to Behavior Algorithms\\for Fighting Games}
\author{Ignacio Gajardo, Felipe Besoain, and Nicolas A. Barriga
\thanks{I. Gajardo, F. Besoain, and N.A. Barriga are with the Escuela de
Ingeniería en Desarrollo de Videojuegos y Realidad Virtual, Facultad de Ingeniería at Universidad de
Talca. Campus Talca, Chile. (e-mail: igajardo16@alumnos.utalca.cl,
fbesoain@utalca.cl, nbarriga@utalca.cl).}
\thanks{Corresponding Author: N.A. Barriga. (e-mail: nbarriga@utalca.cl)}
}
\renewcommand\IEEEkeywordsname{Keywords}
\maketitle
\begin{abstract}
The quality of opponent Artificial Intelligence~(AI) in fighting videogames is crucial. Some other game
genres can rely on their story or visuals, but fighting games are all about the
adversarial experience.
In this paper, we will introduce standard behavior algorithms in videogames,
such as Finite-State Machines and Behavior Trees, as well as more recent
developments, such as Monte-Carlo Tree Search. We will also discuss the existing
and potential combinations of these algorithms, and how they might be used in
fighting games.
Since we are at the financial peak of fighting games, both for casual players
and in tournaments, it is important to build and expand on fighting game
AI, as it is one
of the pillars of this growing market.
\end{abstract}


\IEEEpeerreviewmaketitle
\section{Introducción}
\IEEEPARstart{E}{l} género de peleas en los videojuegos es uno en los que más se ha presenciado
una evolución con el paso del tiempo, con videojuegos contemporáneos del género
rompiendo récords en ventas~\cite{SSBUSales} y reflejada junto a los mismos
videojuegos, gracias a los avances en software y hardware que la industria ha
presentado. Desde la época de los \textit{arcades} este género ha tenido un
enfoque en la parte humana de su diseño, tomando como modo principal su aspecto
multijugador y competitivo, pero ¿qué ocurre si no se tiene un oponente contra
el cuál combatir?

%problema
En este artículo exploraremos múltiples técnicas de Inteligencia 
Artificial~(IA) utilizadas actualmente en videojuegos de pelea, como Máquinas de Estado
Finito~(FSM, Finite-State Machine) o Árboles de
Comportamientos~(BT, Behavior Tree), así como también
algoritmos de uso incipiente, como \emph{Monte-Carlo Tree Search}~(MCTS).
Incluiremos ejemplos de su utilización e implementación, y
características específicas de la IA en juegos de pelea.

%plan del paper
En la sección~\ref{sec:tras} definiremos las características principales de los
videojuegos de peleas y luego explicaremos los fundamentos de los algoritmos
utilizados en el resto del artículo. La sección~\ref{sec:pelea} describe cómo se
integran estas técnicas en los juegos de pelea. Luego la sección~\ref{sec:disc}
discute las ventajas y desventajas de las técnicas presentadas anteriormente, y
explora posibles mejoras. Finalmente concluimos con un resumen y posibles
trabajos futuros.

\section{Trasfondo}\label{sec:tras}

%\subsection{Videojuegos de pelea}
Desde sus primeros momentos de vida, los videojuegos han tomado directa
inspiración de deportes como fútbol, béisbol, tenis, etc. Todo deporte conocido
mundialmente tenía su respectivo videojuego, por lo que era cuestión de tiempo
que las artes marciales recibieran el tratamiento virtual. Juegos como
\emph{Karate Champ} y \emph{Yie Ar Kung-Fu} fueron los primeros que popularizaron
el género y crearon
una plantilla para su futuro, con mecánicas que se mantienen hasta el
día de hoy, como:

\begin{itemize}
    \item Vista lateral en dos dimensiones.
    \item Encuentros del mejor de tres.
    \item Barras de vida como condición de victoria.
    \item Variedad de personajes para elegir con habilidades y estilos de pelea distintos.
\end{itemize}

\begin{figure}[b]
\centering
\includegraphics[width=.9\columnwidth]{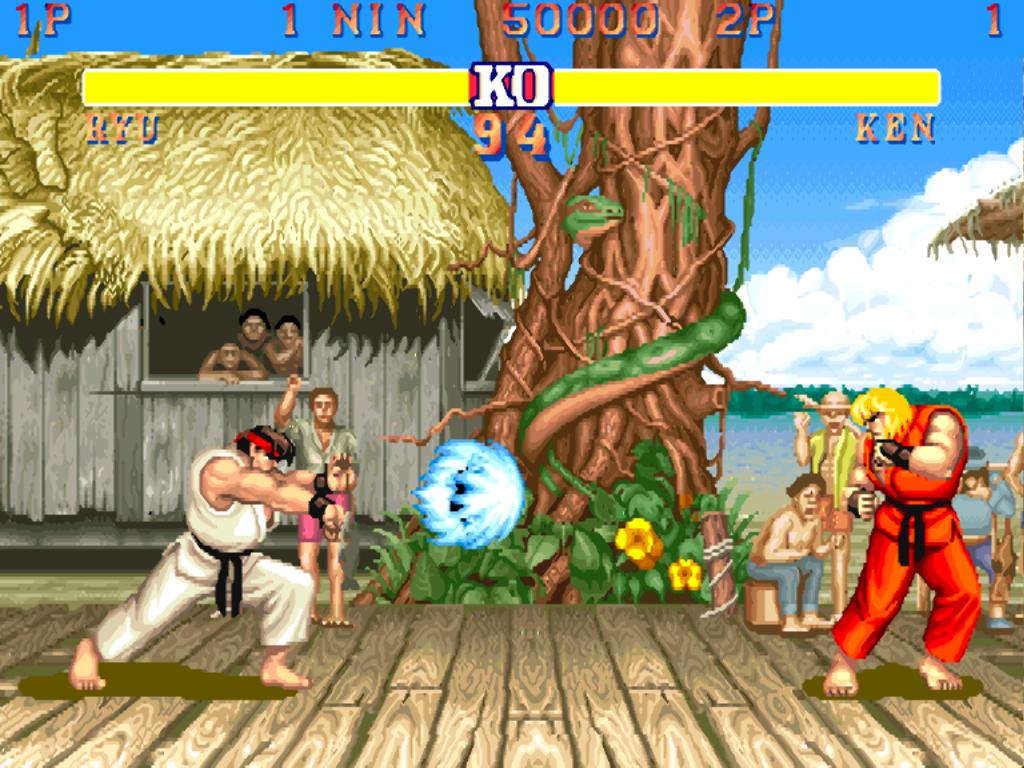}
\caption{Street Fighter II.}
\label{fig:sfii}
\end{figure}

Aunque los juegos mencionados anteriormente fueron ciertamente pilares en el
género, fue \emph{Street Fighter II}~(figura~\ref{fig:sfii}) el que
se popularizó como un
videojuego
competitivo, con un gameplay balanceado, controles de 6 botones, movimientos
especiales y sin duda alguna, el sistema de combos.

\begin{figure}[t]
\centering
\includegraphics[width=\columnwidth]{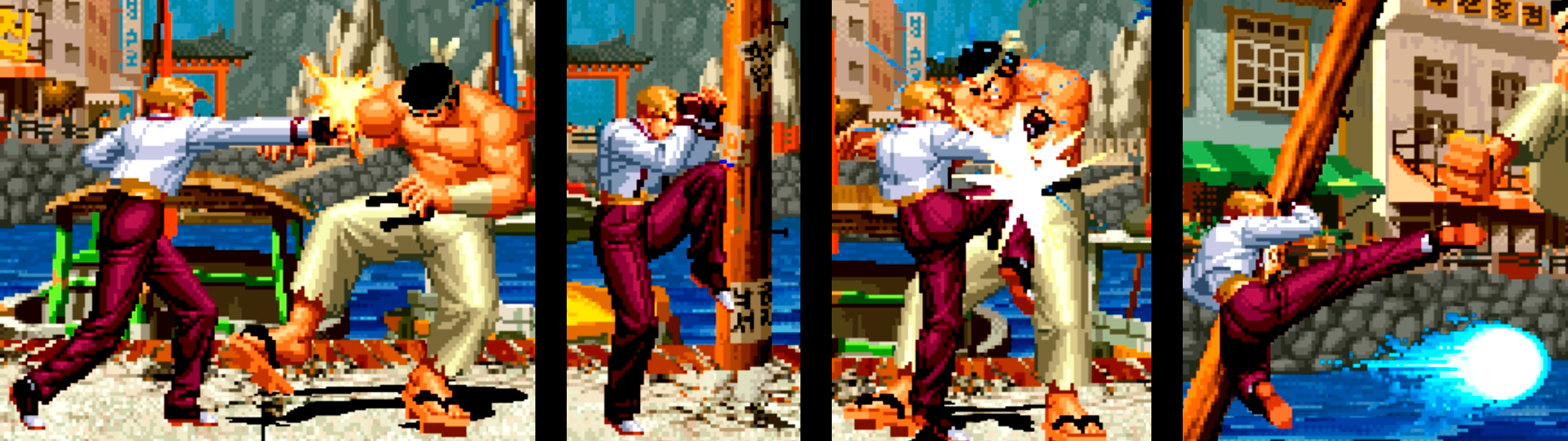}
\caption{Ejemplos de acciones. De izquierda a derecha: ataque, bloqueo, agarre,
movimiento especial.}
\label{fig:moves}
\end{figure}

En su estado más básico los videojuegos de pelea consisten el uso de tres eventos
que el usuario puede ingresar, representando acciones usadas en peleas
reales: ataque, bloqueo y agarre (ver figura~\ref{fig:moves}). Cada
una de éstas “vence” a una y “pierde” contra la otra, funcionando exactamente
como el popular “piedra, papel o tijeras” donde:
\begin{itemize}
\item	Ataque vence a Agarre
\item Agarre vence a Bloquear
\item	Bloquear vence a Ataque
\end{itemize}
Esta trinidad es lo necesario para tener el gameplay fundamental, un set de
opciones que pueden ser vencidas entre ellas. Pero como hemos aprendido con el
paso del tiempo, la complejidad de las batallas ha aumentado progresivamente con
la adición de nuevas mecánicas que agregan más opciones de juego. La mecánica
más relevante y ampliamente utilizada es la de movimientos especiales (ver figura~\ref{fig:moves}). 
Éstos son
variantes del ataque tradicional, que requieren un patrón específico de entradas
para ser ejecutados, pero que tienen propiedades especiales como: lanzar
proyectiles, inmovilizar al enemigo, producir una gran cantidad de daño, etc.
Éstos suelen ser únicos para cada personaje.

\begin{figure*}[ht]
\centering
\includegraphics[width=.9\textwidth]{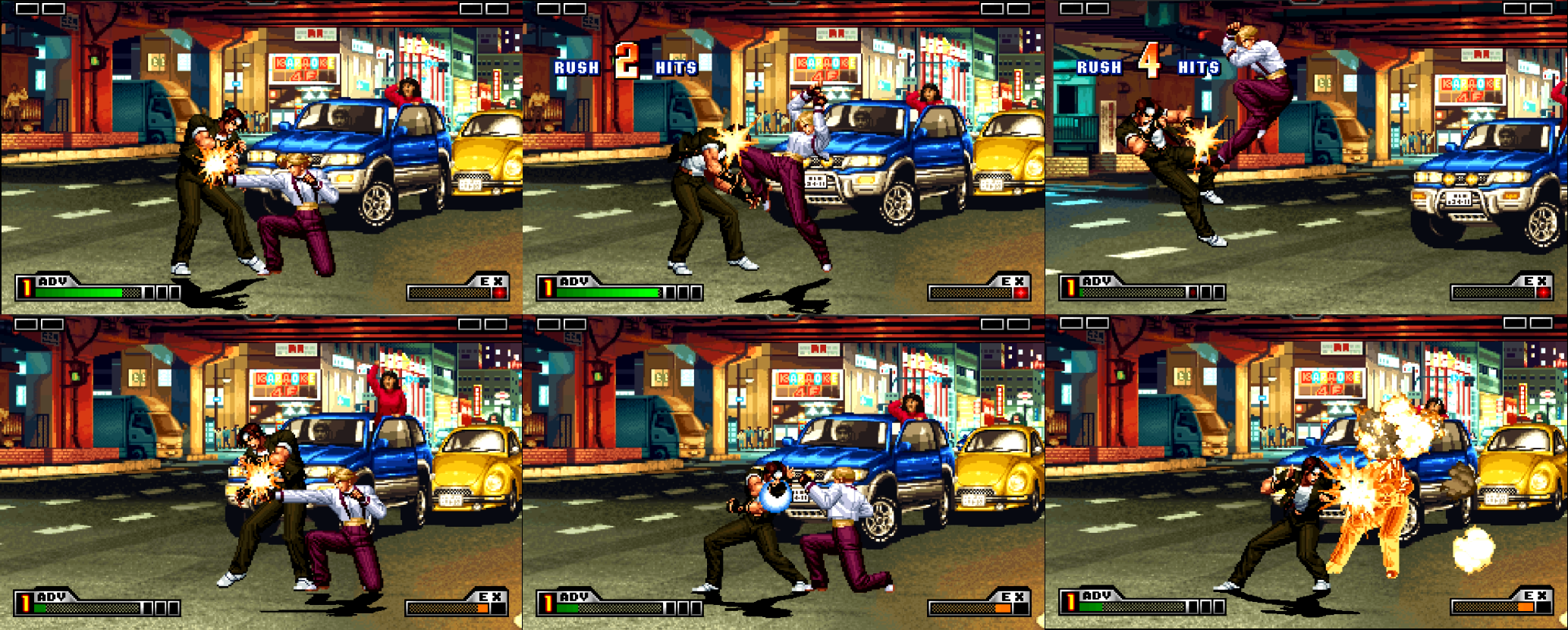}
\caption{Ejemplo de combos. En la secuencia superior se aprecia que todos los
ataques conectan con un resultado positivo, mientras que en la secuencia
inferior el enemigo puede parar el combo y contraatacar.}
\label{fig:combo}
\end{figure*}

Sin embargo, no se puede hablar de los juegos de pelea sin mencionar el sistema
de combos, otra mecánica popularizada por \emph{Street Fighter II}. Un combo se
define como una serie de ataques consecutivos imposibles de bloquear, siempre y
cuando el primer ataque conecte con el enemigo~\cite{combogigantbomb}.  Aunque a
principios del género los combos se mantenían como una mecánica oculta para el
jugador, en la actualidad los juegos cuentan con un sistema donde se les enseña
a aplicar combos creados por los diseñadores, donde se indica la secuencia de entradas y el tiempo
necesario para ser empleado. Usualmente esta opción se encuentra en modos de
entrenamiento, donde se enfrenta a un enemigo que se mantiene estático y se
ignoran las reglas de tiempo y vida (figura~\ref{fig:combo}). 

Los combos funcionan como el alma de cualquier videojuego de pelea, siendo la
principal fuente de
daño en combate y la encargada de demostrar la creatividad y proeza del jugador.

Esta mecánica es desarrollada en la fase de diseño de cada videojuego de pelea
y su naturaleza secuencial permite ser implementada en la IA de un enemigo controlado por el juego, funcionando como una
opción
ofensiva óptima y como una forma de enseñarle cuando y cómo utilizar dicho combo
al jugador.

\subsection{Finite State Machines}
Conociendo la gran cantidad de modelos existentes para la implementación de IA,
una de las primeras y más comunes son las Maquinas de Estados
Finitos~(FSM\footnote{del inglés: Finite-State Machine}), que, gracias a su bajo
requerimiento en conocimientos de programación, referencia visual y estudio
constante, se ha mantenido como un modelo popular para IA en videojuegos (en
especial videojuegos de pelea). 

Las FSM son modelos de computación basados en una máquina hipotética
compuesta por uno o más estados y reglas de transición asociadas. Estos estados
pueden ser accedidos uno a la vez,
con la máquina seleccionando los estados accesibles y revisando las condiciones y consecuencias de moverse a otro estado
por medio de transiciones predefinidas~\cite{bevilacqua2013finite}.

\begin{figure}[t]
\centering
\includegraphics[width=.9\columnwidth]{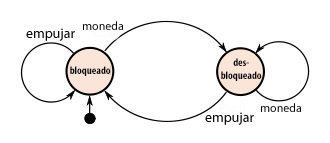}
\caption{Ejemplo de un torniquete.}
\label{fig:torniquete}
\end{figure}

Las FSM siguen una lógica relativamente sencilla, comenzando por un estado inicial, donde el personaje comienza el proceso de “toma de
decisiones”, en el cual considera una serie de posibles transiciones y los estado en los cuales desembocan. Si una de estas condiciones se cumple se pasará al
estado adecuado. En el caso de que ninguna de las condiciones se verifique, se
mantendrá en el estado actual continuamente hasta que se cumpla alguna
condición.

La figura~\ref{fig:torniquete} presenta un simple ejemplo de FSM con el uso de
un torniquete, un dispositivo con varias barras giratorias para que las personas pasen de una en una, a fin de facilitar su control. En su
estado inicial el torniquete se encuentra bloqueado, y tiene 2 posibles
transiciones, empujar la barra de seguridad o ingresar una moneda, lógicamente
podemos inferir que es necesario ingresar dinero al torniquete para su uso por
lo que solamente empujar la barra no cambiará su estado, por otro lado, ingresar
una moneda nos transfiere al estado desbloqueado, que nos entrega las mismas
transiciones en, pero en el contexto opuesto dando como resultado un intercambio
entre los roles de las transiciones. Si el torniquete ya se encuentra
desbloqueado ingresarle más monedas no cambiará su estado, mientras que empujar
la barra de seguridad nos devolverá a su estado bloqueado. 

Como fue mencionado anteriormente, la implementación de FSM y su comprensión es
relativamente sencilla~\cite{graham2004reusable,fu2004ultimate}. Sin embargo, en proyectos de 
mayor envergadura, a medida que las FSM crecen, puede haber un deterioro en la calidad del código, al existir transiciones
que pueden ocurrir en todos los estados. En el ejemplo del
torniquete, si se quiere crear un estado donde el torniquete deje de funcionar,
se tendría que poner el mismo estado con la misma transición en los dos estados
ya existentes, creando una duplicación innecesaria.

Para arreglar estas situaciones existe una variante de las FSM, llamada máquina
de estado finito jerárquica (HFSM, por su sigla en inglés), lo que permite, en
términos simples, crear estados dentro de estados. Así en nuestro ejemplo del
torniquete creamos 2 estados nuevos, funcional y no funcional, dejando todos
nuestros estados previos dentro de funcional, así solo es necesaria una nueva
transición. Gracias a su simple implementación, estructura definida y facilidad
de representarse visualmente lo hace uno de los modelos más populares, en
especial cuando se quiere ejecutar en equipos de trabajo que contengan
integrantes no programadores, como, por ejemplo, en equipos de desarrollo de
videojuegos. 

\subsection{Behavior Trees}

Una alternativa para la creación de IA son los llamados Árboles de
Comportamiento~(BT\footnote{del inglés Behavior Tree}),
que a diferencia de una FSM, no mantienen un estado, si no que toman
\emph{comportamientos} que
la IA puede ejecutar si se cumplen
las condiciones necesarias. Dichos comportamientos de organizan en un árbol
jerárquico 
que controla las decisiones de la IA en cuestión. El árbol 
comienza en la raíz dándole inicio a la búsqueda, desde aquí se puede pasar a 
un nodo. Si éste no es una hoja (nodo sin hijos) éste cumple una función 
compositora, las que pueden ser:
\begin{description}
\item[Secuenciador:] Los secuenciadores, como su nombre lo indica, visita a 
cada uno de sus hijos, del primero al último. En el caso de que alguno falle en su
 ejecución se retorna el \textit{fracaso} al padre, en el caso contrario, si el último hijo
 en la
  secuencia es ejecutado sin problemas, se retorna \textit{éxito}. 
\item[Selector:]  El selector también visita cada nodo hijo en orden, pero la
 condición para pasar de un nodo a otro es que el primero falle, pasando por cada
  uno hasta encontrar el nodo exitoso, retornando \textit{éxito}. Si todos los nodos fallan 
  se retorna \textit{fracaso}.
\end{description} 
Al llegar a un nodo hoja, éste comienza a ser ejecutado, por
lo que retornará \textit{en funcionamiento}, \textit{éxito} o \textit{fracaso} dependiendo de su condición. En 
el caso de que retorne \textit{en funcionamiento} se reproducirá en cada tic hasta que 
retorne \textit{éxito} o \textit{fracaso}. 
Para ejemplificar el nodo hoja tomaremos un proceso de caminar, que consiste
en una función que tiene como parámetros un personaje y un destino. Que el 
programa retorne \textit{en funcionamiento} significa que el personaje está en camino al 
destino, o sea, caminando; si se retorna \textit{éxito} significa que el personaje llegó al 
destino y si retorna \textit{fracaso} significa que el personaje no pudo llegar al destino.

La versatilidad del modelo no sólo permite crear manualmente una inteligencia
artificial competente, si no que también se pueden aprender 
automáticamente~\cite{ABT2019}.

\subsection{Monte-Carlo Tree Search}
A diferencia de los métodos presentados anteriormente, que permiten a un
diseñador especificar comportamientos, Monte-Carlo Tree Search 
(MCTS)~\cite{kocsis2006bandit} es
un algoritmo de búsqueda adversaria para seleccionar los mejores
movimientos posibles. El enfoque principal
de MCTS es crear reproducciones de posibles victorias, comenzando con una
muestra de movimientos aleatorios que llegan hasta el final del juego entregando
valores a los nodos del árbol. Así los nodos más altos son más propensos a ser
usados en próximas iteraciones futuras, por ejemplo, después de cada movimiento
legal del jugador actual.  Cada ronda de búsqueda de un MCTS consiste en 4
pasos: Selección, Expansión, Simulación y Propagación.

\begin{description}
	\item[Selección:] se seleccionan los movimientos desde la raíz, hasta una
  hoja, tomando en cuenta las estadísticas de victoria o derrota de dichas
  acciones en iteraciones anteriores.
\item[Expansión:] si no
se llega al final del juego, sigue la expansión, donde se agregan
nuevos nodos al árbol que equivalen a todos los movimientos válidos. 
\item[Simulación:] se seleccionan movimientos de forma
aleatorizada, simulando el juego hasta el final.
\item[Propagación:] se utiliza el resultado de
la simulación para actualizar las estadísticas de los nodos entre el nodo hoja y
la raíz.
\end{description}

\section{IA en Videojuegos de Pelea} \label{sec:pelea}
%\subsection{Inteligencia Artificial en los videojuegos de pelea} 
La competencia es vital en los videojuegos de pelea. La mayoría de éstos son
creados en base a su faceta multijugador. Por esto, en el mercado contemporáneo
los recursos van directo al balanceo del juego y la creación de nuevos
personajes con estilos de juego nuevos y únicos, que son agregados mediante
parches y DLC\footnote{del inglés: DownLoadable Content}. Estos cambios son
relevantes para los usuarios más \emph{hardcore} (jugadores que planean mantenerse jugando por meses o años y jugar este juego específico pasa a ser parte de su día a día), dependiendo de la longevidad del
videojuego. 

Pero, esto no significa que los modos de un jugador (y en consecuencia, el
desarrollo
de inteligencia artificial) queden de lado, es más, en los últimos años se
ha demostrado que la valoración por estos modos de juego no es menor, especialmente
en el grupo de jugadores casuales. 

Esto ha sido demostrado en dos instancias recientes, la primera siendo la
negativa
recepción de la versión inicial del juego \emph{Street Fighter V} lanzada en el
2016,
al abstenerse de un modo \emph{arcade}, siendo éste el principal método de juego
singleplayer en otros juegos de la saga~\cite{SFVreview}. La otra, es el éxito
obtenido por la saga \emph{Mortal Kombat}, que vió un reinicio con la salida de
\emph{Mortal Kombat (2011)} que resultó ser un éxito comercial y aclamado por la
crítica, especialmente por su modo historia siendo el mayor responsable por su
popularidad en el mercado casual~\cite{mortalkombat}. 

Una de los problemas más grandes al momento de implementar una IA es lo
dinámico que este género tiende a ser, con el jugador pasando a distintos
estados constantemente, cambiando de tácticas ofensivas a defensivas, creando
oportunidades para iniciar un combo, buscando la forma de acabar con la vida del
oponente para luego intentar ganar por un “time out”, y muchas otras más. Es por
esto que antes de tomar una decisión la IA debe considerar muchos factores en el
entorno y la situación en la que se encuentra. Por ejemplo, algunos factores
que afectan las decisiones que la IA debe tomar son:

\begin{itemize}
\item	La distancia entre los dos personajes.
\item	Si el oponente se encuentra atacando.
\item	Si se encuentra un proyectil en la pantalla.
\item	El daño que ha sido infringido al oponente.
\item	Si es que se encuentra de cara a la izquierda o a la derecha.
\item	Si es que se encuentra contra la pared.
\end{itemize}

\subsection{FSM en los Videojuegos de Pelea} 
Como fue
mencionado
anteriormente, los FSM se han mantenido constantemente como una solución viable
para la creación de inteligencia artificial en videojuegos, ya que trae grandes
ventajas a largo plazo en el desarrollo de un proyecto. Por ejemplo, éstas
pueden planearse detalladamente junto a otros miembros del equipo desarrollador
antes de su implementación. Pero más específicamente, se toma en cuenta que la
mayoría de las inteligencias artificiales en los videojuegos son reactivas, es
decir, actúan de forma dependiente al input del usuario.

\begin{figure}[ht]
\centering
\includegraphics[width=.9\columnwidth]{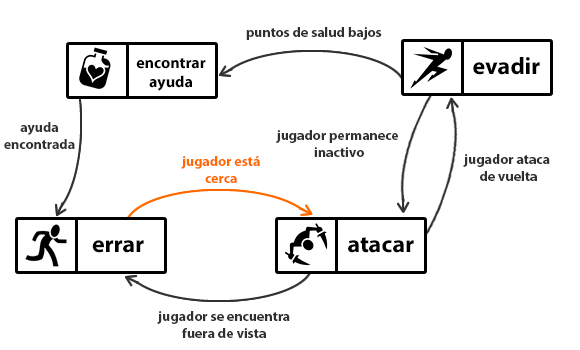}
\caption{Acciones de un enemigo. Adpatado de \cite{bevilacqua2013finite}.}
\label{fig:fsm}
\end{figure}
En la figura~\ref{fig:fsm} se puede apreciar un ejemplo en un contexto de un
enemigo
genérico para un videojuego con un nivel de complejidad mayor al ejemplo del
torniquete.  Aquí se puede apreciar el concepto de inteligencia reactiva, al ser
la mayoría de las transiciones dependientes del mismo estado del jugador, ya sea
si éste es visible al enemigo o si está atacando, además de acciones
exclusivamente respondiendo a peligro en la búsqueda de ayuda al encontrarse
bajo en puntos de vitalidad. A menos que se quiera crear una inteligencia
artificial muy compleja e independiente, utilizar una FSM es probablemente una
buena primera opción.

En el contexto de los juegos de pelea, la alternativa inicial que fue
usada es conocida como \textit{input reading}, donde el juego usa los 
inputs del jugador para determinar los movimientos de sí mismo. Ésta información
también es usada para definir los patrones de acciones de la IA enemiga en
tiempo real. Las principales ventajas de \textit{input reading} son su simple
implementación. Al contar con el sistema de \textit{ataque, bloqueo y agarre}
existe
una
respuesta correcta a cualquier input ingresado, y también facilita su balanceo
de dificultad, determinando la regularidad del uso de esta respuesta correcta,
así cometiendo
más errores en dificultades menores. Sin embargo, con el paso del tiempo se
transformó en una implementación detestable para los jugadores, al hacerse
imposible de combatir en dificultades altas, al siempre conocer lo que el jugador
quiere hacer y contrarrestarlo efectivamente. Si bien dificultades altas e
injustas eran muy comunes en la época de los arcades, la transición de los
juegos de
pelea a consolas de hogar creó una necesidad por un nuevo modelo de IA, y es
aquí donde las FSM entran en escena.

Las máquinas de estado finitas han sido utilizadas en varios videojuegos de
pelea, atrayendo incluso a desarrolladores de franquicias clásicas y juegos
creados por compañías AAA, como 
\emph{Killer Instinct (2013)}~\cite{killerinstinctAI}.

En cuanto a su funcionamiento, la etapa inicial consiste en obtener los datos de
jugadores reales en batallas reales. La mayoría de las veces se buscan a
jugadores de nivel medio-alto y un conocimiento completo de las mecánicas del
juego. Para una recolección de información completa y relevante, se suelen
llevar una gran cantidad de peleas en el mismo contexto, contra los mismos
jugadores, luego de cada iteración de la inteligencia artificial, para asegurar de
que los datos estén completos y actualizados. La primera ventaja que podemos ver
en el uso de
FSM, es cómo el sistema de combos se aplica fácilmente en un entorno de pelea
real,
ya que la IA puede guardar los patrones necesarios para que sea ejecutado,
teniendo como única condición que el primer ataque conecte con el enemigo.

Para la parte más amplia del juego, la IA recolecta dos tipos de información:
la situación en la que se encuentra en base a uno de los parámetros
contenidos en la FSM, y los pasos que llevar a cabo durante la
situación en la que se encuentra. A éstas las llamaremos estado y táctica,
respectivamente. El parámetro evaluado en cuestión dependerá de un valor de
prioridad entregado a cada uno, usualmente teniendo como alta prioridad las
condiciones de victoria (cantidad de vida y tiempo) y como baja prioridad las
acciones que cambien la estrategia en cuestión, como la cantidad de veces que ha
sido realizado cierto movimiento.

Los estados son determinados al programar la inteligencia artificial, revisando
las partidas jugadas por los humanos para tomar en cuenta las situaciones más
cruciales. Las tácticas se determinan al buscar los inputs más
utilizados por los jugadores en los estados evaluados, teniendo así una
agrupación de operaciones a realizar. Es aquí donde la inteligencia artificial
define por su cuenta qué operación es la mejor para lo que se quiere cumplir, ya
que no sólo se quiere saber los movimientos que se utilizaron correctamente por
el jugador humano, sino que también se deben utilizar los errores cometidos ya
que la
intención no es crear una IA imbatible, si no crear una IA que retenga el
interés del jugador.

\subsection{MCTS en Juegos de Pelea}
Su funcionamiento implica el uso de MTCS como herramienta de selección, siendo
responsable de evaluar los distintos requerimientos encontrados en el género de
peleas (vida, posición, tiempo, etc.) para luego crear un árbol en base a las
posibles soluciones, entregarles un valor de que representa la viabilidad de
cada opción, para luego seleccionar el que considere eficiente para la situación
presentada~\cite{MCTSFG2016}. Si bien con esto se tiene lo necesario para crear
una IA para videojuego de pelea, MTCS puede no ser la mejor opción, ya que
al contar
con una gran cantidad de movimientos y, en particular, situaciones en donde se
aplican estos movimientos, se le impide tomar decisiones en un tiempo razonable.
Es
por esto que a continuación se presentará una de las formas para disminuir el
tamaño de la búsqueda y, por consecuencia, su tiempo.

\subsubsection{Combinar MCTS con FSM}
La unión de máquinas de estado finito con Monte-Carlo Tree Search permite crear
una IA más flexible y fuerte. Existen dos posibles implementaciones
para esta
unión. 
La primera y más común consiste en el uso de FSM de forma estándar, donde
el luchador contará con una cantidad definida de estados en base al tipo de
acción que se quiere realizar y transiciones que definen los requerimientos para
este estado. Asemismo, estos estados tendrán un grupo de movimientos asociados
a la acción que se quiere realizar, como un \textit{pool} de distintas opciones
unidas por
el mismo propósito. Luego de seleccionar el tipo de estado a pasar, entra MCTS
para la selección de movimiento(s) a realizar, donde usa un método de selección
estocástica, analizando un árbol con exploradores como UCB1 para encontrar el
movimiento (o secuencia de movimientos) más eficiente. 

Por ejemplo, una FSM que contenga un estado de acciones ofensivas y otro de
acciones defensivas toma la información que el juego le entrega y transiciona
al primer estado. Luego, en vez de decidir que acción ejecutar, le entrega una
lista de acciones posibles (ej.:un uppercut o gancho de boxeo, una patada en el aire, una
secuencia de combos) al MCTS el cual decide la mejor jugada.

La implementación alternativa propone un enfoque opuesto a la versión presentada
anteriormente. La FSM mantiene el funcionamiento de estados con movimientos y
transiciones, pero MCTS toma el rol de seleccionar a que estado se debe pasar.
Es decir, MCTS realiza su búsqueda sobre una serie de acciones abstractas, las
transiciones entre estados.

\section{Discusión} \label{sec:disc}
A continuación, presentamos algunos comentarios sobre las técnicas
mencionadas anteriormente y examinamos posibles vías de desarrollo, utilizando
comparaciones y propuestas alternativas para resolver el objetivo analizado.

\subsection{BT en Juegos de Pelea}
Behavior Trees en videojuegos de pelea es un método poco explorado, pero, al
tener un funcionamiento jerárquico se puede crear una inteligencia artificial
competente. Al contener naturalmente una jerarquía en forma de árbol, con
el uso de selectores y secuenciadores, se puede adoctrinar a un personaje a hacer
ciertas acciones en cierto contexto.  Con las condiciones de juego ya definidas,
los primeros nodos a implementar serían los selectores, al tener el trabajo de
decidir el estilo de juego a tomar, en base a la situación en la que se encuentre
el personaje. Por ejemplo, depender más de una estrategia defensiva o
arriesgarse
por un combo.  Por otra parte, los secuenciadores se agregarían a la IA al
haber situaciones que requieran más de una acción. Esto puede ser
necesario no solamente en combos, el cual obviamente requiere más de una acción
para ser funcional, y puede ser cancelado si en algún momento del combo un golpe
no conecta, sino que también sirven como una forma de “reiniciar” el escenario
en el
que se encuentra, para volver a una situación neutral o llevar a cabo una acción
defensiva inmediata en el caso de que se realice un movimiento muy arriesgado. 

\subsection{FSM vs Behavior Trees} 
Ahora que tenemos un conocimiento de cómo funcionan ambos modelos, podemos
compararlos efectivamente para entender las ventajas y desventajas de cada uno.

Por el lado de las FSM, se mantiene la ventaja anteriormente mencionada de ser
un
modelo muy visual, incluso por sobre BT (con su representación de árbol),
dándole
una simpleza y elegancia que otros modelos carecen y agilizando los
procesos
de diseño y ejecución, haciéndola particularmente atractiva en implementaciones
pequeñas o medianas. Además, este modelo se ha mantenido relevante por un largo
tiempo, por lo que existen muchos ejemplos para aprender y mejorar su uso.

Como desventajas, el uso de FSM se puede complicar si se implementa una gran
cantidad de estados, donde las condiciones se hacen muy específicas y las
transiciones se enredan entre sí. Asimismo, este problema puede ocurrir al no
ser posible para una FSM ejecutar dos o más estados simultáneamente, requiriendo
crear más de una FSM, lo cual no es ideal, ya que perdería la simpleza que antes
la caracterizaba~\cite{FSMProsCons}.

En contraste, los BT cuentan con la característica de ser modulares, es decir,
el árbol
creado puede ser estudiado, diseñado y trabajado en partes separadas como
subárboles. Esta es una ventaja que destaca especialmente cuando
se trabaja con sistemas de mayor tamaño y complejidad. Adicionalmente, la
organización jerárquica de los árboles permite mantener un mayor orden en las
acciones de cada nodo y sus hijos, permitiendo un análisis más cómodo y
controlado. 

Una de las desventajas de BT es que para asegurar su uso óptimo se debe trabajar
de forma paralela y optar por una lectura que funcione horizontalmente y
verticalmente. También, aunque BT sea considerado un modelo rápido, su uso se
puede transformar en una desventaja si la lectura del árbol se tiende a caer en
loops, aumentando el costo de cada ejecución y perjudicando la
búsqueda~\cite{BTProsCons}.

\subsection{Combinar MCTS con BTs}

Monte-Carlo Tree Search aplicado en behavior trees tiene los mismos principios
que cualquier BT en juegos de pelea, con selectores para decidir las acciones a
tomar y secuenciadores para llevar a cabo una sucesión de acciones relacionadas
entre sí, pero, agregando MCTS esta propuesta puede ser más inteligente. 

El uso de MCTS tiene paralelos con la primera implementación presentada en FSM,
donde MCTS decide los movimientos específicos luego de entrar al estado deseado,
en BT luego de pasar por el selector el árbol pasa a un nodo que contiene un
MCTS donde hace su trabajo de escoger movimientos.  

Implementar un MCTS no solo le daría una mayor efectividad al árbol, 
acortaría el tamaño de los secuenciadores al tomar su trabajo en ciertas
decisiones. Mientras que  BT aporta al acelerar el usualmente lento
procesamiento de MCTS (al disminuir las acciones que éste debe considerar),
encontrando un balance entre velocidad y calidad de la acción seleccionada.

\section{Conclusión}
Gracias a que el género de videojuegos se ha mantenido relevante con el paso del
tiempo, especialmente en estos últimos años, las investigaciones en el tema
seguirán ampliando los horizontes en la búsqueda de la inteligencia artificial
“perfecta”, una que encuentre el punto medio entre una dificultad entretenida
para un jugador casual y una desafiante para un jugador competitivo, sin
sacrificar la credibilidad de una inteligencia que se sienta humana. Aunque
hasta ahora FSM tiende a ser el modelo primordial y popular, se ha visto un
interés en otros modelos, como el ya mencionado Behavior Tree, que tienden
a una inteligencia más adaptable y cercana al comportamiento humano. Éste
siempre ha sido el objetivo en cuanto a la creación de inteligencias
artificiales en juegos de pelea.

\balance

Además, el interés general en el tema se ha masificado a un público más casual y
joven por medio de competencias como \emph{FightingGameAIC}~\cite{lu2013fighting},
donde los participantes crean sus propios combatientes para competir por quien
puede crear una inteligencia artificial imbatible, poniendo a pelear sus
inteligencias artificiales entre ellas para decidir cuál es la que mejor se
adapta a distintas situaciones y obtiene resultados ganadores.
Es en esta misma competencia donde se han encontrado nuevas implementaciones de
inteligencias artificiales innovadoras con la unión y mejora de distintos
modelos, y es en competencias como ésta donde se puede probar el modelo
propuesto de Behavior Tree junto con Monte-Carlo Tree Search en un ambiente
adecuado.

Se puede apreciar que el futuro se ve prometedor, demostrando que distintos
modelos ofrecen nuevas formas de implementar una IA competidora y este número
sigue expandiendo~\cite{lu2013fighting}, ya que con el conocimiento que se ha obtenido en los últimos
años, esta investigación podría llegar a modelos como Utility Systems y demás,
que contienen el potencial necesario para considerar su evaluación.

\bibliographystyle{IEEEtran}
\bibliography{refs}

\end{document}